\documentclass[a4paper,conference]{IEEEtran}
\IEEEoverridecommandlockouts

\usepackage{cite}
\usepackage{amsmath,amssymb,amsfonts}
\usepackage{algorithmic}
\usepackage{amsmath}
\usepackage{graphicx}
\usepackage{textcomp}
\usepackage{xcolor}
\usepackage{hyperref}
\usepackage{multirow}

\def\BibTeX{{\rm B\kern-.05em{\sc i\kern-.025em b}\kern-.08em
    T\kern-.1667em\lower.7ex\hbox{E}\kern-.125emX}}
\begin{document}

\title{HiMFR: A Hybrid Masked Face Recognition Through Face Inpainting}

\author{\IEEEauthorblockN{Md Imran Hosen$^1$, Md Baharul Islam$^{1,2*}$}
\textit{$^1$imranjucse21@gmail.com, $^2$bislam.eng@gmail.com,}\\
\textit{$^1$Department of Computer Engineering, Bahcesehir University, Istanbul, Turkey}\\
\textit{$^2$College of Data Science \& Engineering, American University of Malta, Bormla, Malta}
}

\maketitle

\begin{abstract}
To recognize the masked face, one of the possible solutions could be to restore the occluded part of the face first and then apply the face recognition method. Inspired by the recent image inpainting methods, we propose an end-to-end hybrid masked face recognition system, namely HiMFR, consisting of three significant parts: masked face detector, face inpainting, and face recognition. The masked face detector module applies a pretrained Vision Transformer (ViT\_b32) to detect whether faces are covered with masked or not. The inpainting module uses a fine-tune image inpainting model based on a Generative Adversarial Network (GAN) to restore faces. Finally, the hybrid face recognition module based on ViT with an EfficientNetB3 backbone recognizes the faces. We have implemented and evaluated our proposed method on four different publicly available datasets: CelebA, SSDMNV2, MAFA, {Pubfig83} with our locally collected small dataset, namely Face5. Comprehensive experimental results show the efficacy of the proposed HiMFR method with competitive performance. Code is available at \url{https://github.com/mdhosen/HiMFR}
\end{abstract}


\section{Introduction}
Face recognition has gained a lot of attraction and is still one of the most prominent research fields in computer vision and pattern recognition. It has become a trend for its practical and commercial applications, including face attendance system, face access control,  mobile payment through face authenticity, and so on \cite{wang2020masked}. Many works have been reported while deep learning-based work \cite{liu2017sphereface,geitgey2019face,deng2019arcface,song2019occlusion, hariri2021efficient,montero2021boosting,junayed2021deep} with the efficient loss function has shown promising recognition results. Weiyang Liu et al., \cite{liu2017sphereface} proposed SphereFace model\textcolor{red}{:} angular-based softmax loss function, which allows convolution neural network (CNN) to learn angularly discriminative features. However, many approximations are required to compute this loss that causes unstable network training. To solve training unstability, Jiankang et al.,  \cite{deng2019arcface} proposed ArcFace (Additive Angular Margin Loss) model to improve the discriminative ability of the model and stabilize the training process.

A significant face portion, including the nose and mouth, is covered with a mask that shortens the description parts of the face. Thus, the extension of existing face recognition methods for masked face recognition can significantly reduce the performance. According to a recent National Institute of Standards and Technology (NIST) \cite{257486}, face recognition system performance drop nearly fifty percent when faces cover with a mask, and false negatives increase. In contrast, false positives were unchanged or slightly reduced \cite{dharanesh2021post}. Several unconventional approaches have recently been introduced for masked face recognition with different techniques such as discarded occlusion parts \cite{song2019occlusion,hariri2021efficient,junayed2021deep} and combining training (both mask and unmask) \cite{montero2021boosting}.

To recognize the masked face through discarding the mask part, the first mask portion is discarded and then combined with the applicable dictionary (set up by taking advantage of the distinctions between the top conv elements of hidden and visible face pairs) items. It is multiplied with the proper features to neglect the features that come from the masked part in recognition \cite{song2019occlusion,hariri2021efficient,junayed2021deep}. This technique's advantage is that it needs less time to train since it works with comparatively fewer data. { However, this method provides poor results for mixing applications (masked and unmasked ) due to partial visibility, which leads to a drop in information while dealing with unmasked faces. For example, while \cite{junayed2021deep} got 95.07\% accuracy for masked face, their performance reduced to 91.10\% for mixture application.
In \cite{montero2021boosting} throughout the training system, both masked and unmasked datasets are shuffled one by one with the usage of the identical seed, and they decide the newly selected face image coming from the masked or unmasked data with 50\% probability. However, this technique achieves good accuracy but requires a lot of training data with high computational costs}.

Additionally, it is necessary to identify the masked face accurately and timely to recognize a masked face. Though some CNN based methods \cite{nagrath2021ssdmnv2,ahmed2021,sethi2021face} have achieved competitive performance. There are some challenges to recognizing masked faces, including (i) the necessity for a highly accurate and efficient masked face detector, ii) partial visibility that reduces the description of the face, and iii) the necessity for a robust and computationally efficient face recognition model. To address the challenges mentioned earlier, we integrated three major modules: masked face detection, occlusion restoration or inpainting based on Generative Adversarial Network (GAN), and finally, face recognition using a hybrid Vision Transformer (Vision Transformer with EfficientNetB3 \cite{tan2019efficientnet} backbone).
The contributions of this paper are summarized as follows:

\begin{figure*}[htb]
\centerline{\includegraphics[width=0.95\textwidth]{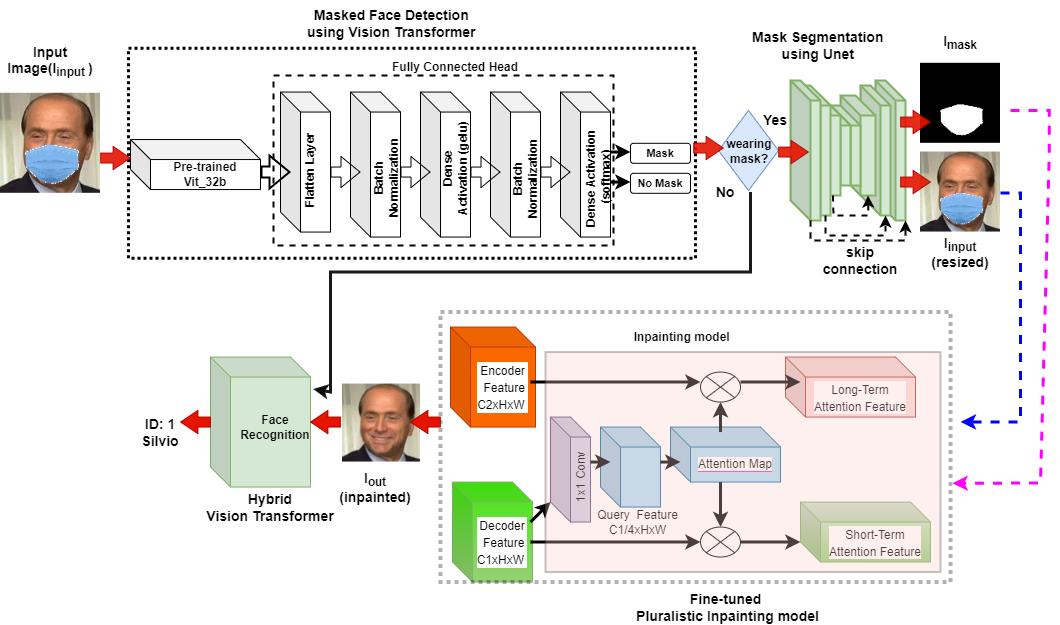}}
\vspace{-0.2cm}
\caption{The overall architecture of the proposed hybrid face recognition model that performs in three major modules: (i) masked face detector which determines face covered with mask or not, (ii) face restoring or inpainting module that is applied to restore the face, if the face is covered with mask and (iii) face recognition module.}
\label{methodologhy}
\end{figure*}

\begin{itemize}
    \item A hybrid face recognition module based on the Vision Transformer integrated with a masked face detector and a face inpainting module has been proposed. Our HiMFR can automatically detect and remove the mask and restore the occlusion part to recognize the person.

    \item We propose a masked face detection module using the ViT\_b32 with a fully connected (FC) head that provides competitive performance compared to different pre-trained models.

    \item To make the restoration faces convenient and realistic, we have fine-tuned the Pluralistic Image Completion (PIC) \cite{zheng2019pluralistic} method that can restore a large face occlusion region with more than one possible outcome.
    
    \item Our proposed HiMFR method is tested with four publicly available dataset including CelebA \cite{liu2015faceattributes}, SSDMNV2 \cite{nagrath2021ssdmnv2}, MAFA \cite{mafa}, Pubfig83 \cite{kumar2009attribute} and our collected Face5 dataset. The experimental results show promising performance to detect and recognized masked face compared to the state-of-the-art methods.
\end{itemize}

\section{Proposed HiMFR Method}
Our HiMFR model can be divided into masked face detection, face inpainting, and face recognition. Firstly, the masked face detection module {determines whether the image contains a face mask or not}. The intermediary face inpainting module is performed to restore the face before the face recognition module. The overall architecture of the proposed HiMFR model shows in Fig. \ref{methodologhy}.

\begin{figure*}[htb]
\centerline{
\includegraphics[width=0.92\textwidth]{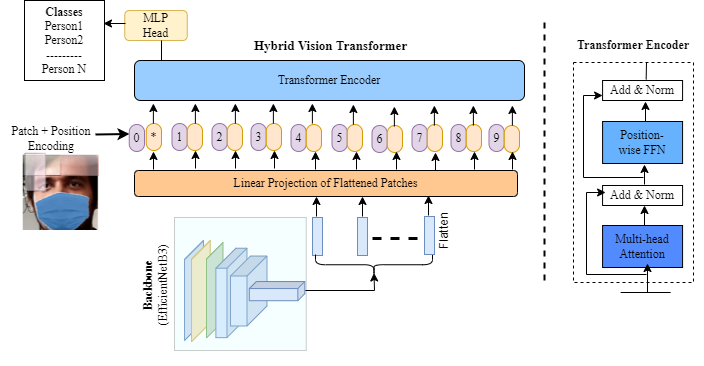}}
\vspace{-0.6cm}
\caption{The overall architecture of our proposed hybrid face recognition module is based on the ViT transformer and EfficientNetB3 as a backbone. Token has been generated with sliding patches rather than overlapping to increase performance. The encoder part is a standard Transformer encoder. }
\label{vitarchitecture}
\end{figure*}

\subsection{Masked Face Detector}
Our mask face detector module takes $I_{input}$ as input, and { provides a decision whether the mask is present or not in the face. The difference between a traditional face detector and our masked face detector is that the face detector determines the face portion. In contrast, our mask face detector provides a binary decision about mask detection}. To reduce the computational cost and make the train faster, we have used transfer learning where pre-trained ViT\_b32 has been used as a feature extractor. It trained on ImageNet21k \cite{deng2009imagenet} dataset which consists of around 14 million images. In the fine-tuning process, the last layer (fully connected head) of the ViT has been chopped, and a fully connected new head has been added to the pre-trained ViT\_b32. The fully connected layer follows up on a packed input where all neurons are associated with each input. The FC head consists of a flattened, two batch normalization layers and two dense layers as shown in Fig. \ref{methodologhy}. We have used Categorical Cross-Entropy (CE) and Rectified Adam as loss functions and optimizers. The Categorical Cross-Entropy loss function shown in Eq. \ref{lossfunction} is used in multi-class classification tasks where one predictable output belongs to one of many possible categories. Additionally, this function is intended to measure the distinction between two likelihood distributions.
\begin{equation}\label{lossfunction}
   CE = \sum_{n=1}^{n} y_i \log \hat{y_i}
\end{equation}
where $n$ is the number of output size. $\hat{y_i}$ is the predictable output of $i^{th}$ target value of $i^{th}$ sample. 

\subsection{Face Inpainting}
\begin{figure}[ht!]
\centering
\centering
\begin{minipage}{1\textwidth}
\hspace{0.5cm} Input \hspace{0.6cm} Completion 1 \hspace{0.1cm} Completion 2 \hspace{0.05cm} Completion 3
\end{minipage}
\includegraphics[width=0.115\textwidth]{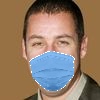}
\includegraphics[width=0.115\textwidth]{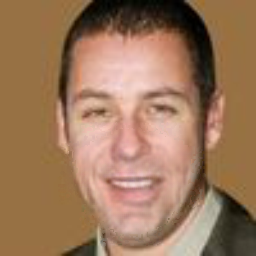}
\includegraphics[width=0.115\textwidth]{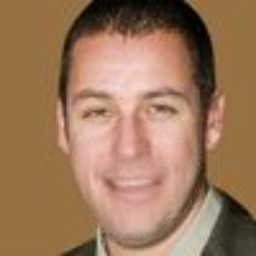}
\includegraphics[width=0.115\textwidth]{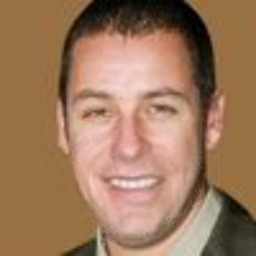} \vspace{0.1cm}

\includegraphics[width=0.115\textwidth]{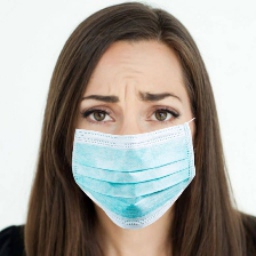}
\includegraphics[width=0.115\textwidth]{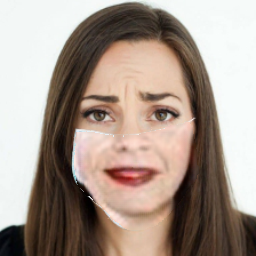}
\includegraphics[width=0.115\textwidth]{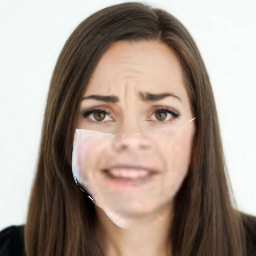}
\includegraphics[width=0.115\textwidth]{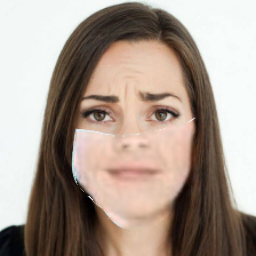}

\vspace{-0.2cm}
{\caption{\label{diverseOutput} \footnotesize The diverse and plausible outputs from the single input. First input from CelebA \cite{liu2015faceattributes} and second input from real world MAFA \cite{mafa} dataset. }}
\end{figure}

We have fine-tuned the pluralistic image completion (PIC) model initially proposed by Zheng et al. \cite{zheng2019pluralistic}  which can produce more than one plausible outcome. {The benefit of this model is that one can verify the person's identity in possible ways}. Assume there is an original image ($I_g$) whose quality has been degraded due to some missing pixels. This degradation is represented by $I_m$  (the masked partial image). $I_c$ is also defined as the ground truth hidden pixels' complement partial image. The image completion method generates the missing pixels in a deterministic way. As a result, it is possible to reconstruct only one image. However, the authors used a non-deterministic way to generate multiple possible images in the PIC method \cite{zheng2019pluralistic}. {Though the original PIC model can create various images for each person, we have developed a single image based on their discriminator scores in our approach. Additionally, one can verify the person with diverse possible outputs shown in Fig. \ref{diverseOutput}. The three possible image completion results are shown by wearing the synthetic and real-world masks}. The PIC model needs the binary mask to perform image generations. To segment the mask from the face, we used a pre-trained segmentation model \cite{din2020novel}.

\subsection{Face Recognition} 
To recognize faces after occlusion removal, we have implemented a hybrid ViT that was initially proposed by Dosovitskiy et al.\cite{dosovitskiy2020image} with EfficientNet \cite{tan2019efficientnet}  that employs a revolutionary version scaling strategy. It uses a simple but powerful compound coefficient to scale up CNN's more structured manner. Traditional techniques use freely scaled network dimensions such as width, intensity, and resolution, while EfficientNet scales each variable uniformly using a predetermined set of scaling factors. We chopped the last layer of EfficientNet and added it (without the head) to the ViT. We have frozen the EfficientNet weights and implemented the sliding patches technique \cite{zhong2021face}  to overlap the patches to increase the accuracy. Our model follows the ViT architecture shown in Fig. \ref{vitarchitecture}.

The overall architecture of the Vision Transformer model is as follows: (i) creating patches from an image (fixed sizes), (ii) making the image patches flat, (iii) creating lower-dimensional linear embeddings, (iv) adding positional embeddings, (v) feeding the sequence into the transformer encoder as an input, (vi) image labels use to train the ViT model, which is fully supervised shown in Eq.\ref{vit1} and Eq.\ref{vit2}. The first step is the ViT separates an image into square patches in a grid and flattens the patch into a single vector. After that, learnable position embeddings enable the model to understand the image structure. In our implementation, hyperparameters such as learning rate, number of heads, transformer layer, etc., have been set. Then, we augmented our data, and Multi-Layer Perceptron (MLP) was defined. The two-dimensional neighborhood structure is only utilized once in the model to divide the picture into patches at the model starting and to fine-tune the position embedding for images of varying resolutions \cite{dosovitskiy2020image}.

\begin{equation}\label{vit1}
 z_0 = [Y_{class};Y^1_pE;Y^2_pE;.....;Y^N_pE] + E_{pos},  
\end{equation}
\begin{equation} \label{vit2}
\begin{split}
z\prime_l & =  MSA (LN(z_{l-1})), l=1...L\\
z_l & =  MLP (LN(z\prime_l)) + z\prime_l  , l=1...L\\
y & =  LN(z^0_L)
\end{split}
\end{equation}
where, $Y_p$ represents flattened patches, $Y_pE$ indicates output patch embedding and $Y_{class}$ = $z_0$ is a learnable embedding.  MSA stands for multiheaded self-attention, and LN is Layer-norm. $z_0$ serves as the image representation of $y$.

After defining MLP, we have implemented patch creation and patch encoding layer. By projecting a patch into a vector of size projection dimension, the Patch Encoder layer linearly converts it. It also adds to the projected vector a learnable position embedding. Our ViT model comprises 2 Transformer blocks, each of which uses the 8 Multi-Head Attention layer as a self-attention mechanism for the patch sequence. The Transformer blocks generate a tensor with a projection dimension, batch size, and several patches, then processed via a classifier head with a softmax activation function to generate the final probabilistic result.

\section{Dataset and Experiment}
\subsection{Dataset}
To evaluate the efficiency of the HiFMR method and compare with state-of-the-art methods, we conduct experiment with four publicly available datasets, including CelebA\cite{liu2015faceattributes}, SSDMNV2\cite{nagrath2021ssdmnv2}, MAFA \cite{mafa}, Pubfig83 \cite{kumar2009attribute} and our locally collected Face5 dataset. Data has been pre-possessed (e.g., cropping) to achieve better performance from each module. For instance, to train the image inpainting module, it needs to be higher quality with the proper size. Besides that, the image should contain only the face, so cropping is also necessary. The image in CelebA dataset \cite{liu2015faceattributes} cropped into size $256 \times 256$ of the face. Then we create simulated mask  \cite{anwar2020masked}.


CelebA \cite{liu2015faceattributes} dataset contains 200K celebrity images which have large quantities, diversities, and rich annotations. Two other publicly available datasets, e.g., SSDMNV2\cite{nagrath2021ssdmnv2} contains around 11k data (5500 are mask and others are unmasked face), and MAFA \cite{mafa} contains 6k data (3k with mask and 3k unmask face) have been used to compare the masked face detection module with state-of-the-art methods. We collect 5k data (size $224 \times 224$) from 5 individuals focusing on the face. We follow the data collection procedure by providing consent to the subjects. We have used a 2 MP camera resolution frame rate at 30 (frames per second) with the size of $480 \times 640$ pixels. { We have also used Pubfig83 \cite{kumar2009attribute} (consist of 83 classes and each class contains around 100 images) dataset to train and testing of our face recognition module}.

\subsection{Experimental Setup}
Experiments are conducted in windows 10, 32GB RAM, and one GPU (NVIDIA Geforce RTX 2070). To train and test each module, we split the dataset into 80\% and 20\% for training and testing. To train the masked face detector module, we have used processed CelebA unmask and synthetically created mask data (size $224 \times 224$). {The term synthetically indicates that the data initially was unmasked faces, and then we put the same size mask artificially on the faces}. We have run the experiment for five epochs and batch size $16$ with Rectified Adam optimizer and a $0.0001$ learning rate. We have run the image inpainting module for $150$ epochs where Adam optimizer has been used with a learning rate of $0.0001$ and  $256 \times 256$ image size. We have used our Face5 dataset for face recognition module training and run it for 10 epochs with num\_heads 4, batch size 2, transformer layers 2, num\_heads 8, image size $224\times224$. Additionally, the Adam optimizer has been used with a fixed learning rate of $0.0003$.

\subsection{Evaluation Matrix}
All modules of the proposed HiMFR method are separately evaluated. Accuracy, Precision, Recall, and F1 Score are calculated  shown in Eq. \ref{prcn} - \ref{f2scr}.
\begin{equation}\label{prcn}
    Precision = \frac{TP}{(TP+FP)}
\end{equation}
\begin{equation}\label{rcl}
    Recall = \frac{TP}{(TP+FN)}
\end{equation}
\begin{equation}\label{acrcy}
    Accuracy = \frac{TP+TN}{(TP+FP)+(TN+FN)}
\end{equation}
\begin{equation}\label{f2scr}
 F1Score =2*\frac{Precision*Recall}{Precision+Recall}
 \end{equation}
where $TP$, $TN$, $FP$ and $FN$ indicate true positive, true negative, false positive, and false negative respectively.


{Besides, we have used Structural Similarity Index (SSIM) and Peak Signal to Noise Ratio (PSNR) to compare image inpainting results. The SSIM measures the similarity between two images by comparing three measurements between images: luminance, contrast, and structure. Higher SSIM (up to 1) indicates the output image maintains its original structure. On the other hand, PSNR computes the correlation between the obtained output and the input. A higher PSNR value indicates better quality in the performance.}

\begin{table}[htb]
\caption{Comparison our masked face detection module with different pre-trained models on CelebA \cite{liu2015faceattributes} dataset.}

\vspace{-0.4cm}
\begin{center}
\scalebox{0.94}{
\begin{tabular}{|c|c|c|c|c|c|}
\hline
\textbf{Architecture} & Resnet50 & MobileNetV2 & VGG16 & VGG19 & HiMFR\\
\hline
\textbf{Accuracy}&{99.03\%}&{97.04\%}&{96.56\%}&{96.85\%}&99.90\%\\
\hline
\end{tabular}}
\label{pretrained}
\end{center}
\end{table}
\begin{table}[htb]
\caption{Comparison the performance of the HiMFR masked face detector with state-of-the art methods on two datasets.}
\begin{center}
\scalebox{0.99}{
\begin{tabular}{|c|c|c|c|}
\hline
\multirow{2}{5em}{\textbf{Methods}}  & \multicolumn{2}{|c|}{\textbf{Dataset}} & \multirow{2}{5em} {\textbf{Accuracy}} \\
& MAFA \cite{mafa} & SSDMNV2\cite{nagrath2021ssdmnv2} &\\
\hline
Sethi et al., \cite{sethi2021face}& \checkmark & \text{\sffamily X} & 98.27\% \\
Ahmed et al.,\cite{ahmed2021}& \checkmark & \text{\sffamily X} &93.94\% \\
Nagrath et al., \cite{nagrath2021ssdmnv2}& \text{\sffamily X} & \checkmark & 92.64\% \\
Our HiMFR & \checkmark & \text{\sffamily X} &\textbf{99.04}\% \\
Our HiMFR & \text{\sffamily X} & \checkmark &\textbf{99.32}\% \\
\hline
\end{tabular}}
\label{SAM}
\end{center}
\end{table}


\begin{table}[htb]
\centering
\caption{ Quantitative Comparison of HiMFR inpainting module with state-of-the-art methods in terms of PSNR and SSIM metric.}
\label{quantative}
\scalebox{1}{
\begin{tabular}{|p{2.5cm}||p{1.1cm}|p{1.1cm}|}
\hline
\textbf{Method} & \textbf{PSNR}               & \textbf{SSIM} \\ \hline
Yu et al.,\cite{yu2018generative}                  & 32.15       & 0.74          \\ \hline
 Zhen et al., \cite{zheng2019pluralistic}               & 33.34        & 0.79         \\ \hline
 Din et al., \cite{din2020novel}     & 32.67       & 0.78         \\ \hline
Our HiFMR           & \textbf{35.92}        & \textbf{0.90}         \\ \hline

\end{tabular}}
\label{inpaintQuant}
\end{table}

\begin{figure*}[thb]
\centering
\begin{minipage}{1\textwidth}
\hspace{0.9cm}Input \hspace{1.9cm} G.T. \hspace{1.4cm} Jiahui et al. \cite{yu2018generative} \hspace{0.6cm}Zheng et al.,\cite{zheng2019pluralistic} \hspace{0.5cm}Din et al. \cite{din2020novel} \hspace{0.5cm}Pluralistic (fine-tune)
\end{minipage}
\includegraphics[width=0.15\textwidth]{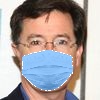}
\includegraphics[width=0.15\textwidth]{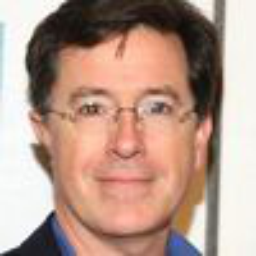}
\includegraphics[width=0.15\textwidth]{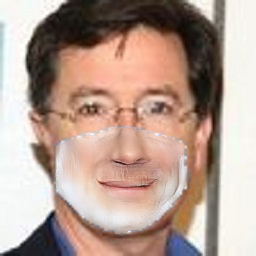}
\includegraphics[width=0.15\textwidth]{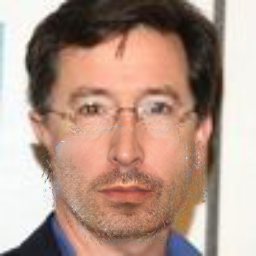}
\includegraphics[width=0.15\textwidth]{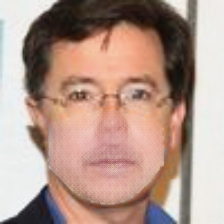}
\includegraphics[width=0.15\textwidth]{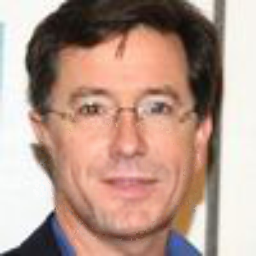}

\vspace{0.05cm}

\includegraphics[width=0.15\textwidth]{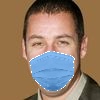}
\includegraphics[width=0.15\textwidth]{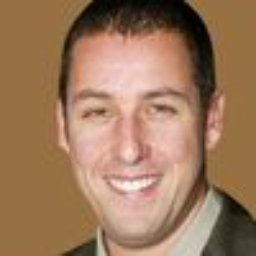}
\includegraphics[width=0.15\textwidth]{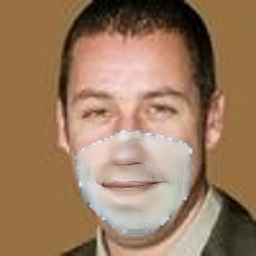}
\includegraphics[width=0.15\textwidth]{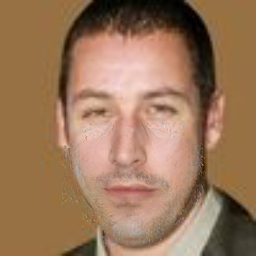}
\includegraphics[width=0.15\textwidth]{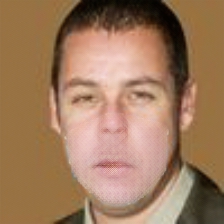}
\includegraphics[width=0.15\textwidth]{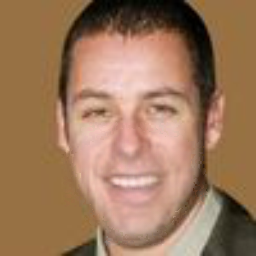}

\vspace{0.05cm}

\includegraphics[width=0.15\textwidth]{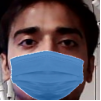}
\includegraphics[width=0.15\textwidth]{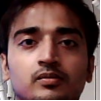}
\includegraphics[width=0.15\textwidth]{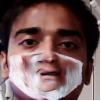}
\includegraphics[width=0.15\textwidth]{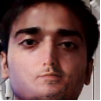}
\includegraphics[width=0.15\textwidth]{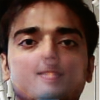}
\includegraphics[width=0.15\textwidth]{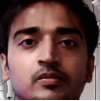}

\vspace{0.05cm}

\includegraphics[width=0.15\textwidth]{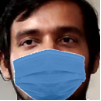}
\includegraphics[width=0.15\textwidth]{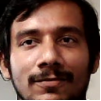}
\includegraphics[width=0.15\textwidth]{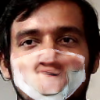}
\includegraphics[width=0.15\textwidth]{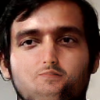}
\includegraphics[width=0.15\textwidth]{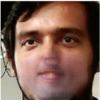}
\includegraphics[width=0.15\textwidth]{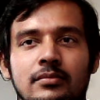}

\vspace{-0.3cm}

\caption{ \small Qualitative comparison of HiMFR fine-tune inpainting module with state of the art methods: on CelebA \cite{liu2015faceattributes} (top 2 rows) and Face5 (bottom 2 rows) datasets.}
\label{inpaintingQuality}
\end{figure*}

\section{Experimental Results}
\subsection{Masked Face Detection Performance}
The HiMFR masked face detector module is time-efficient and provides state-of-the-art performance with the minimum number of iterations shown in TABLE \ref{pretrained}. The Resnet50 achieved comparatively higher accuracy than the other pre-trained models (MobileNetV2, VGG16, and VGG19) in {5280} iterations, while our module achieved slightly higher accuracy within {1760} iterations. It indicates the faster performance of our HiMFR face detection module. To make a fair comparison, we have compared the HiMFR masked face detector module with the state-of-the-art methods demonstrated in TABLE \ref{SAM}. Our module achieves higher accuracy compared to state-of-the-art techniques. Sethi et al., \cite{sethi2021face} acquired 98.27\% accuracy near to our method. 

\subsection{Face Inpainting Performance}
The quantitative evaluation of pluralistic image completion is challenging since it provides various possible solutions for a single masked image. 
However, to make the comparison, {we took the single output from plausible generated outputs based on their discriminator score}. We have conducted the comparison on the mixed dataset (CelebA \cite{liu2015faceattributes} and Face5) in terms of Peak Signal to Noise Ratio (PSNR) and Structural Similarity Index (SSIM) as shown in TABLE \ref{inpaintQuant}. Fine-tune pluralistic inpainting achieved better performance than the original pluralistic \cite{zheng2019pluralistic} and other state-of-the-art methods \cite{yu2018generative, din2020novel}. Additionally, we qualitatively compare the fine-tuned inpainting results with state-of-the-art methods, as presented in Fig. \ref{inpaintingQuality}. Pluralistic \cite{zheng2019pluralistic} method generated outcome better than the \cite{yu2018generative,din2020novel} but fine-tuned pluralistic image completion generates a more realistic output with fine details and structure. {Moreover, to test the applicability of our module in the real world, we have tested our module on the real-world masked MAFA dataset \cite{mafa} shown in Fig. \ref{diverseOutput} (bottom row). As shown, our module produces realistic output. However, a small portion of masked is present. It was due to improper segmentation mask from the face.} 

\subsection{Face Recognition Performance }

\begin{figure}[htb]
\centerline{\includegraphics[width=0.55\textwidth]{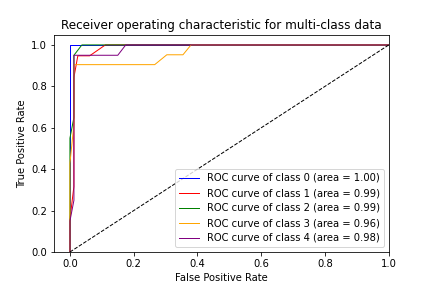}}
\caption{{Receiver operating characteristic (ROC) curve for presenting the relationship between the False and True Positive Rates.}}
\label{rocCurve}
\end{figure}

To demonstrate the performance of our face recognition module, we have trained our face recognition module {on Face5 dataset (five classes: each class contains 1000 images of each individual). The Receiver Operating Characteristic (ROC) curve is plotted using a testing set which consists of 200 images (100 masked faces and 100 unmasked), as shown in Fig. \ref{rocCurve}. {Each color represents the different classes and the area under the curve (AUC) are 1.00, 0.99,0.99,0.96 and 0.98, indicating satisfactory performance}. The performance matrix of our proposed masked face recognition system is demonstrated in TABLE \ref{faceRecognitionMetrices} {for Face5 dataset (five classes and each class represents a individual person (0 $\sim 4$))}. HiMFR achieved a competitive performance with {95\%} accuracy.}

{Additionally, to compare the performance of HiMFR with state-of-the-art methods, we adopted four \cite{liu2017sphereface,geitgey2019face,deng2019arcface,junayed2021deep} deep learning-based face recognition methods. We tested on the Face5 dataset both unmask and mixed (mask and unmask) as shown in TABLE \ref{faceReognitionComparision}. The performance of the state-of-the-art methods dropped significantly for the mixed data, while our masked face recognition shows the robust performance. Besides, our module achieved 94.7\% accuracy on Pubfig83 dataset \cite{kumar2009attribute}, which shows the competitive performance of our module compared to the Savchenko et al.,\cite{savchenko2018efficient} on the same dataset. To see the effectiveness of the backbone, we have trained and tested it without backbone and noticed the face recognize module performance drops significantly.}


\begin{table}[htb]
\caption{Performance matrices for proposed HiMFR on {Face5 dataset.}}
\begin{center}
\scalebox{1}{
\begin{tabular}{|p{2.0cm}||p{1.1cm}|p{1.1cm}|p{1.1cm}|}
\hline
{\textbf{Class label}}&\textbf{Precision}&\textbf{Recall}&\textbf{F-score}\\
\hline
0 & 1.00 & 1.00 & 1.00\\

1 & 0.90 & 0.94 & 0.92\\

2 & 0.95 & 0.95 & 0.95\\

3 & 0.95 & 0.90 & 0.92\\

4 & 0.95 & 0.95 & 0.95\\
\hline
{\textbf{Accuracy} }& \multicolumn{3}{|c|}{{\textbf{0.95}}} \\
\hline
macro avg & 0.95 & 0.95 & 0.94\\
weighted avg & 0.95 & 0.95 & 0.95\\
\hline
\end{tabular}}
\label{faceRecognitionMetrices}
\end{center}
\end{table}

\begin{table}[htb]
\caption{ Comparison of our masked face recognition module with state-of-the-art methods on {Face5 dataset}.}
\begin{center}
\begin{tabular}{|c|c|c|}
\hline
\textbf{Model}&\textbf{Accuracy (Unmask)} & \parbox[t]{3cm}{\centering \textbf{Accuracy (masked and umasked)}}\\
\hline
Liu et al., \cite{liu2017sphereface}& 97.00\% & 81.00 \% \\
Geitgey et al.,\cite{geitgey2019face}& 99.00\% & 83.00\%\\
Deng et al.,\cite{deng2019arcface}& 99.00\% & 90.00\%\\
Junayed et al.,\cite{junayed2021deep}&- & 91.10\%\\
HiMFR&\textbf{99.00}\% & {\textbf{95.00}}\% \\
\hline
\end{tabular}
\label{faceReognitionComparision}
\end{center}
\end{table}



\begin{figure}[htb!]
\centering
\includegraphics[width=0.155\textwidth]{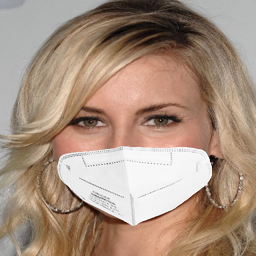}
\includegraphics[width=0.155\textwidth]{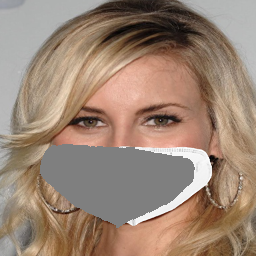}
\includegraphics[width=0.155\textwidth]{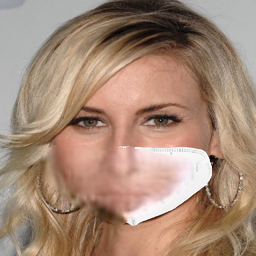} 


\vspace{-0.3cm}
{\caption{\label{failure} \small An example of inaccurate face completion. }}
\end{figure}

\section{Conclusion}
This paper introduced a novel hybrid face recognition through the face reconstruction that works in three sub-modules: masked face detector, face inpainting, and face recognition. A Vision Transformer performed masked face detection. We used pre-trained ViT\_b32 as a feature extractor with a fully connected head that achieved higher accuracy with the minimum iterations and made the masked face detector module time-efficient. The pluralistic image completion model was fine-tuned and used to reconstruct the hidden part of the face covered. Finally, we implemented a Vision Transformer with an EfficientNetB3 backbone-based face recognition module. Our method was evaluated quantitatively and qualitatively with different datasets. 
{Although our HiMFR can recognize masked faces with high accuracy, it can be failed for inaccurate face completion by the image inpainting module (due to imprecise face mask tracking and segmentation). An example of the failure case shown in Fig.\ref{failure} inaccurately. Our method is faster but does not perform in real-time. These limitations may consider in future works.}

\noindent \textbf{Acknowledgements.} This work is supported by the Scientific and Technological Research Council of Turkey (TUBITAK) 2232 Leading Researchers Program, Project No. 118C301.


\bibliographystyle{./IEEEtran}
\bibliography{./IEEEabrv,./IEEEexample}
\end{document}